\documentclass[10pt,twocolumn,letterpaper]{article}

\usepackage{iccv}
\usepackage{times}
\usepackage{epsfig}
\usepackage{graphicx}
\usepackage{amsmath}
\usepackage{amssymb}

\usepackage{csquotes} 
\usepackage[style=ieee, backend=bibtex, mincitenames=1, maxcitenames=2]{biblatex}
\usepackage{comment}
\usepackage{enumitem}
\usepackage{gensymb}
\usepackage{float} 
\usepackage{multirow}
\usepackage{caption}
\usepackage{subcaption}

\newcommand\parsection[1]{\noindent\textbf{#1:}}

\usepackage[breaklinks=true,bookmarks=false]{hyperref}

\iccvfinalcopy 

\def\iccvPaperID{10} 

\ificcvfinal\fi

\bibliography{Sources}

\begin{document}

\title{You can have your ensemble and run it too -- Deep Ensembles Spread Over Time  
}

\author{
\textbf{Isak Meding*}\\
Zenseact\\
\and
\textbf{Alexander Bodin*}\\
Zenseact\\
\and
\textbf{Adam Tonderski}\\
Zenseact\\
Lund University\\
\and
\textbf{Joakim Johnander}\\
Zenseact\\
Linköping University\\
\and
\textbf{Christoffer Petersson}\\
Zenseact\\
Chalmers University of Technology\\
\and
\textbf{Lennart Svensson}\\
Chalmers University of Technology\\
}

\maketitle
\ificcvfinal\thispagestyle{empty}\fi

\begin{abstract}
Ensembles of independently trained deep neural networks yield uncertainty estimates that rival Bayesian networks in performance. They also offer sizable improvements in terms of predictive performance over single models. However, deep ensembles are not commonly used in environments with limited computational budget -- such as autonomous driving -- since the complexity grows linearly with the number of ensemble members. An important observation that can be made for robotics applications, such as autonomous driving, is that data is typically sequential. For instance, when an object is to be recognized, an autonomous vehicle typically observes a sequence of images, rather than a single image. This raises the question, could the deep ensemble be spread over time?

In this work, we propose and analyze Deep Ensembles Spread Over Time (DESOT). The idea is to apply only a single ensemble member to each data point in the sequence, and fuse the predictions over a sequence of data points. We implement and experiment with DESOT for traffic sign classification, where sequences of tracked image patches are to be classified. We find that DESOT obtains the benefits of deep ensembles, in terms of predictive and uncertainty estimation performance, while avoiding the added computational cost. Moreover, DESOT is simple to implement and does not require sequences during training. Finally, we find that DESOT, like deep ensembles, outperform single models for out-of-distribution detection.

\end{abstract}

\section{Introduction}

\begin{figure*}[t]
    \centering
    \begin{subfigure}{0.18\textwidth}
        \centering
        \includegraphics[width=0.95\linewidth]{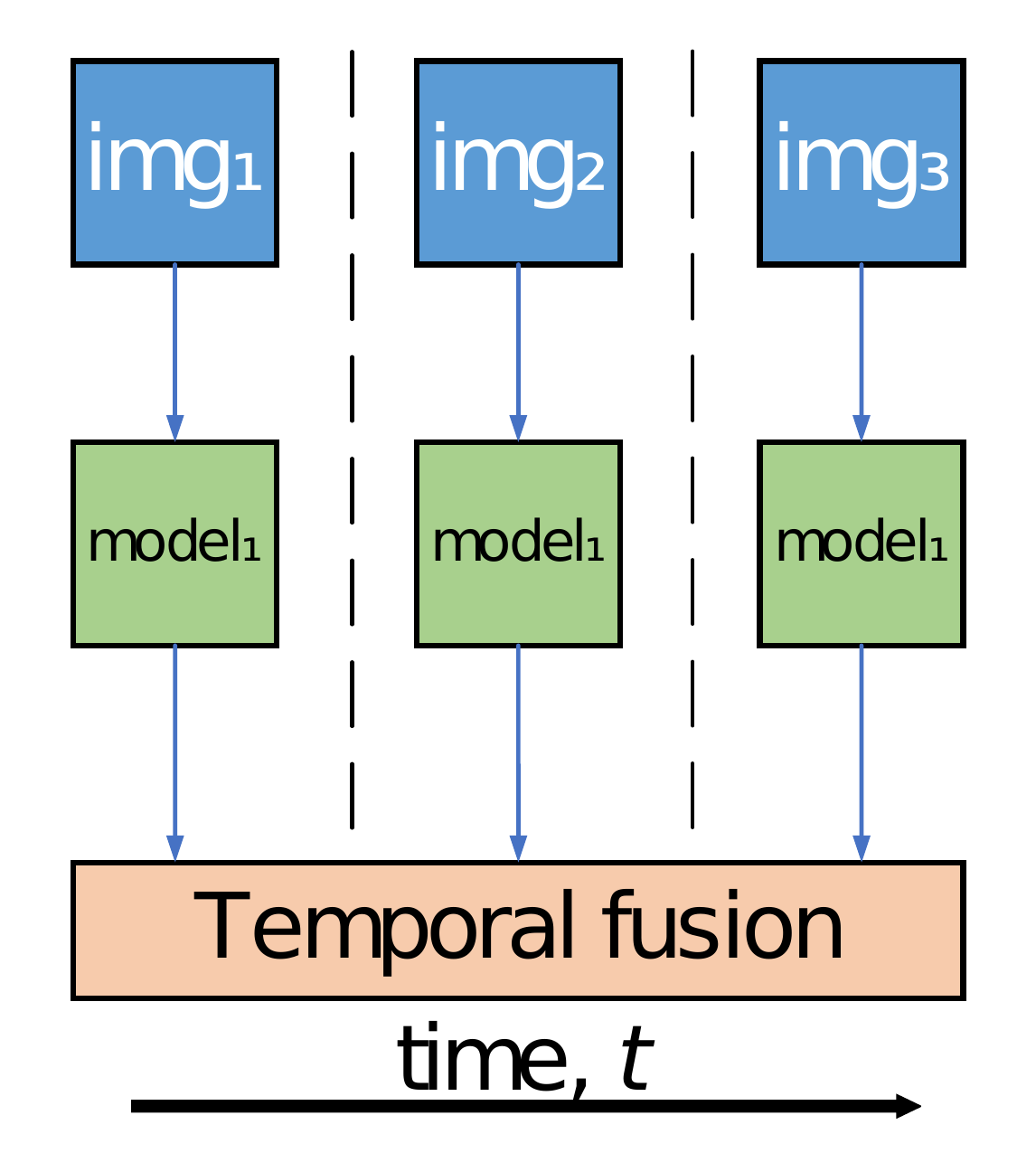}
        \captionsetup{width=\hsize}%
        \caption{Single model.}
        \label{fig:traditional_singlemodel}
    \end{subfigure}
    \begin{subfigure}{0.4\textwidth}
        \centering
        \includegraphics[width=0.95\linewidth]{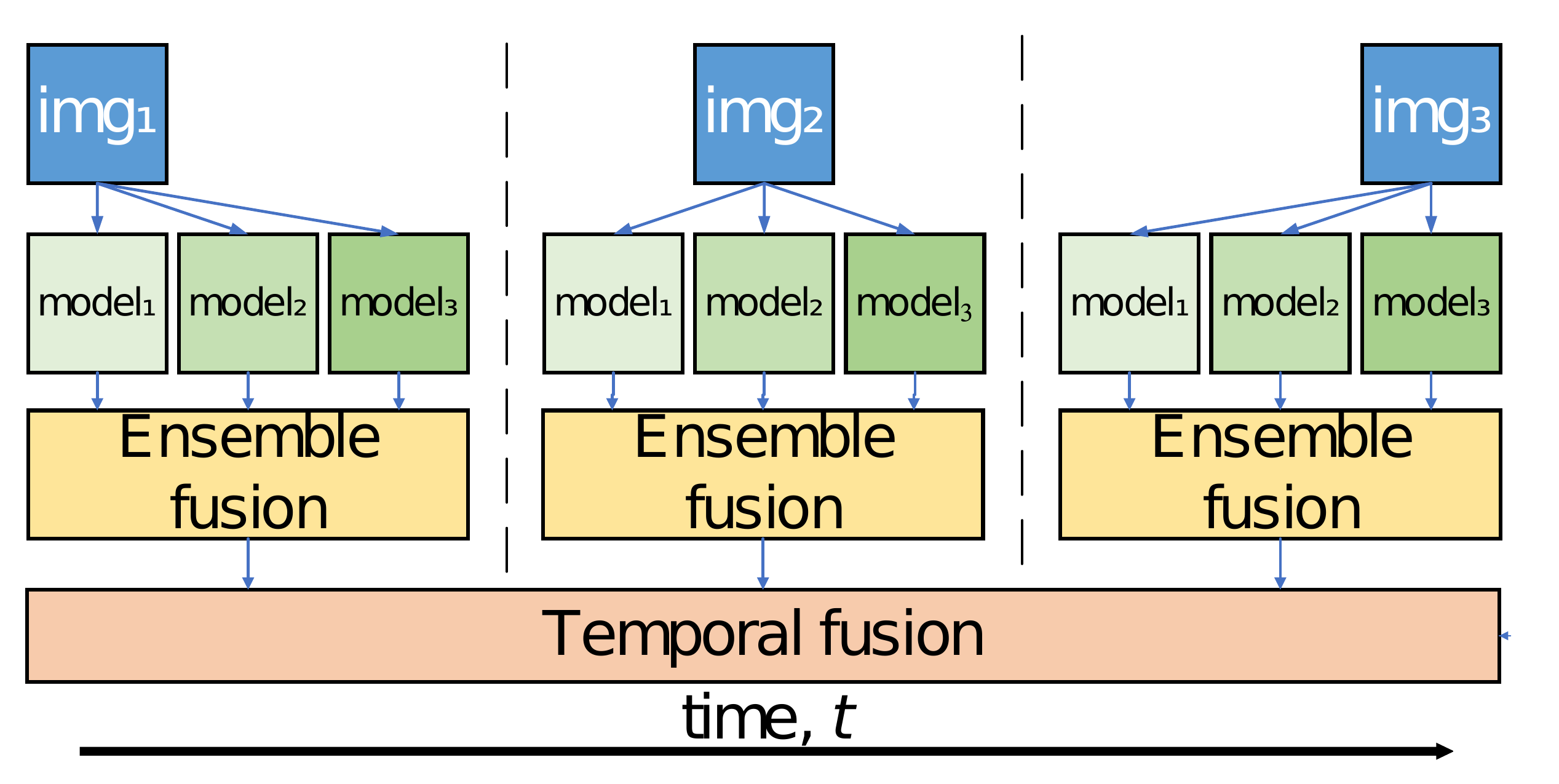}
        \captionsetup{width=\hsize}%
        \caption{Deep ensemble.}
        \label{fig:traditional_ensemble}
    \end{subfigure}
    \begin{subfigure}{0.4\textwidth}
        \centering
        \includegraphics[width=0.95\linewidth]{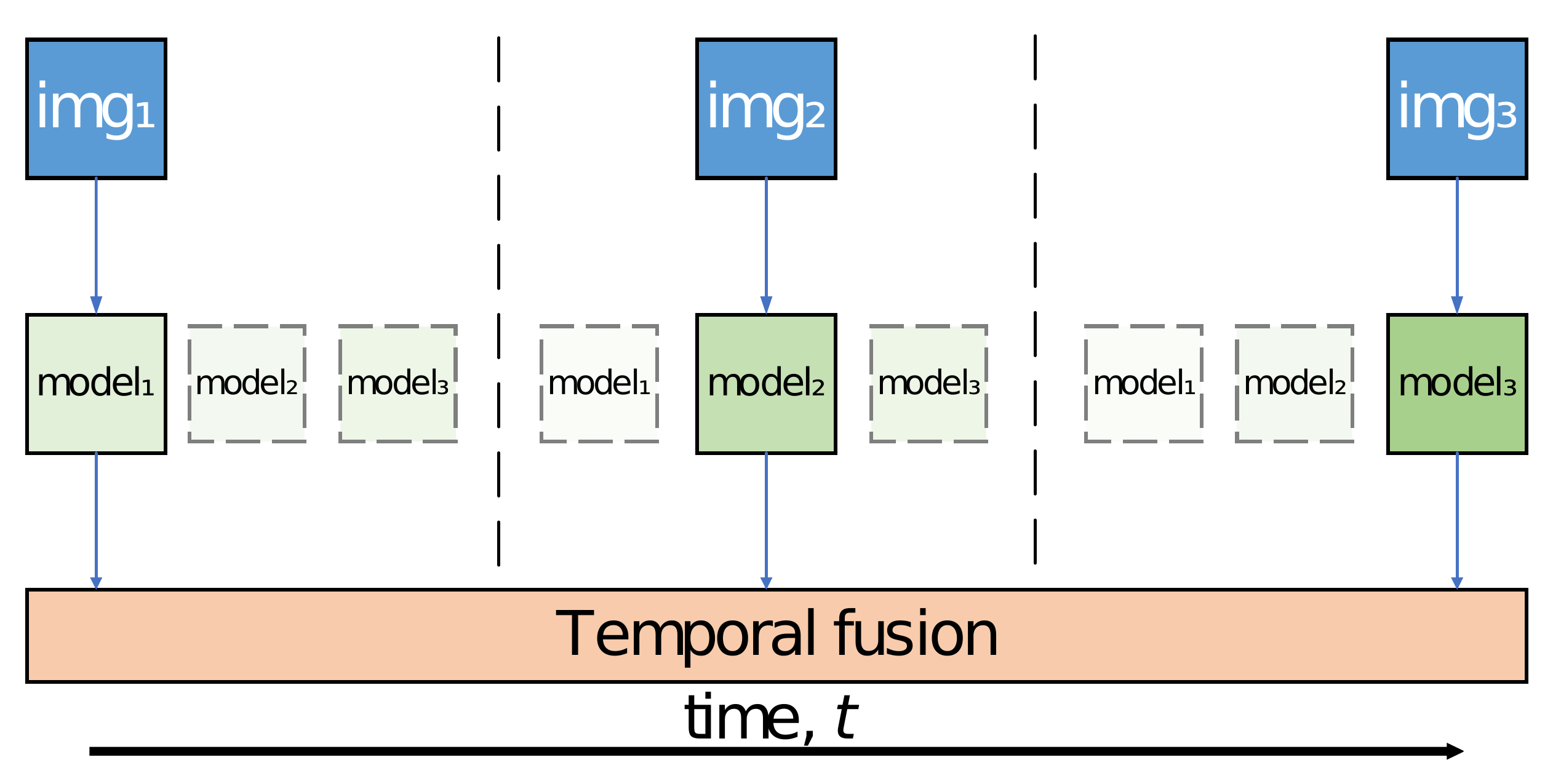}
        \captionsetup{width=\hsize}%
        \caption{Deep ensemble spread over time.}
        \label{fig:DESOT_visualization}
    \end{subfigure}
    \caption{Visualization of the different strategies in an example with three time steps and three ensemble members. The temporal fusion blocks combine predictions produced at each time step using some combination rule. At time step $t$, $t\in \{1,2,...,T\}$ with $T=3$ in this example, the temporal fusion block produces a prediction based on the predictions at time steps $1$ through $t$. The ensemble fusion block combines individual ensemble member predictions into a final ensemble prediction at each time step.}
    \label{fig:model_visualizations}
\end{figure*}

In safety-critical applications, such as autonomous driving (AD), both the predictive performance and the uncertainty quantification performance of neural network models are essential \cite{Amodei:2016, feng2021review, Guo:2017, minderer2021revisiting, liu2020energy}. For example, the posterior probabilities produced by a perception model are required to contain all the information needed for the downstream decision-making systems to make safe and efficient driving decisions. Even for a seemingly mundane task such as traffic sign classification one can imagine the problem that could result from a model misidentifying a speed limit sign for a stop sign on the highway, while nevertheless outputting a high confidence.  
The general problem of overconfidence for certain types of neural networks \cite{Guo:2017} is particularly troublesome in safety-critical applications. 

When it comes to modeling uncertainty, it is often divided into two types of uncertainty -- aleatoric uncertainty and epistemic uncertainty \cite{Kiureghian:2009}. Aleatoric uncertainty is due to inherent randomness in the process that generates the data. It is therefore not possible to decrease this type of uncertainty by improving the model. On the other hand, epistemic uncertainty is that which is caused by a lack of data or knowledge about the underlying process \cite{Abdar:2021}. One example is to predict the outcome of a biased dice throw. It is impossible to predict what side the dice will land on for certain, no matter how accurate the model is. This constitutes irreducible aleatoric uncertainty in the prediction. However, more data can help us identify the bias of the dice and improve the model, thereby decreasing the epistemic uncertainty. A larger number of models reduces epistemic uncertainty, while a larger number of frames reduces aleatoric uncertainty.

Non-Bayesian neural networks often display great predictive performance, but struggle with generating high-quality epistemic uncertainty estimates \cite{Guo:2017,Ovadia:2019,Ashukha:2020}. Bayesian neural networks on the other hand often yield high-quality epistemic uncertainty estimates but are difficult to train \cite{Lakshminarayanan:2017}. One way to improve on the epistemic uncertainty estimation of regular neural networks while improving predictive performance is to use ensembles \cite{Ovadia:2019, Ashukha:2020}. It has long been known that ensembles of neural networks can quantify uncertainty in their predictions~\cite{Muhlbaier:2005} and increase the predictive performance compared to single models \cite{Hansen:1990, Dietterich:2000}. \textcite{Lakshminarayanan:2017} show that neural network ensembles can produce uncertainty estimates that outperform Bayesian models, while also achieving high predictive performance. This allows for a practical and high-performance alternative to Bayesian methods. This type of ensemble is commonly referred to as Deep Ensembles (DEs) \cite{Mehrtash:2020, Mukhoti:2021, Ashukha:2020}, and has been verified by other authors to produce high-quality uncertainty estimates \cite{Ovadia:2019, Ashukha:2020}. 

As previously mentioned, ensembles typically also improve predictive performance over single models. This is mathematically proven in the classification setting by \textcite{Hansen:1990} under the assumption of independent classification errors between ensemble members. They show that if each model achieves an accuracy greater than 50\%, adding more models results in perfect classification performance in the limit. The assumption of independent classification errors does not typically hold in the real world, but the intuition is valid. There are also other potential advantages of using ensembles. \textcite{Wasay:2021} show empirically that for a set model parameter budget, using ensembles results in shorter training times and higher accuracy compared to single models. 

Because of all the benefits of DEs outlined above, applications of them in safety-critical systems, such as AD, would clearly be desirable. However, an important limiting factor for AD applications is the requirement for low latency, real-time processing on resource-limited embedded hardware -- a fact that is typically incompatible with the linear compute increase of DEs in the number of members. 

In this paper, we leverage the fact that the sensor data in AD systems are in general sequential in nature and propose Deep Ensembles Spread Over Time (DESOTs). Instead of applying all ensemble members to the sensor data at each time step, as would be the procedure for a conventional DE, {\it DESOT only applies one member at each time step, but different members at different consecutive time steps}. Hence, DESOT uses the same number of computations as a single model, but the same amount of memory as a DE with an equal number of members. We illustrate the DESOT method by extensively studying the task of traffic sign classification with sequences of tracked patches containing the traffic signs. We choose this task due to its importance in AD applications and its simple formulation, but the method is applicable to a wide range of settings and tasks that use sequence data. We show that DESOTs are competitive with traditional DEs in our setting, both in predictive performance and uncertainty quantification performance. We also show that they outperform both single models and MC-dropout models.

\bigskip
In summary, our contributions are the following:

\noindent\textbf{(i)} We propose Deep Ensembles Spread Over Time (DESOT), an approach applicable to sequences that brings the benefits of Deep Ensembles (DE) without the additional computational cost.

\noindent\textbf{(ii)} We thoroughly analyze the proposed approach on traffic sign classification, where a sequence of tracked patches are fed as input to the model. DESOT obtains the benefit of DEs at the computational cost of a single model. 

\noindent\textbf{(iii)} We thoroughly analyze the out-of-distribution detection performance, based on the entropy of the predictions, and find that DESOT, like deep ensembles, substantially outperforms a single model.

\section{Related Work}

\parsection{Deep ensembles}
Deep ensembles have been shown to be superior to any other ensemble method in uncertainty quantification given a fixed computational budget \cite{Ashukha:2020}. \textcite{Ovadia:2019} also showed that DEs are some of the best-performing models in uncertainty estimation. The intuition behind why this might be is that ensembling is a variance reduction technique, and therefore useful for increasing the quality of epistemic uncertainty estimates \cite{Ueda:1996}.  Furthermore, they have additional advantages in their simple implementation and high predictive performance.

Though deep ensembles have been used for decades, \textcite{Lakshminarayanan:2017} show that they produce state-of-the-art uncertainty estimates. DEs are simple to train, with three basic steps involved: (i) ensure that a proper scoring rule is used as loss function; (ii) optionally use adversarial training to increase robustness; and (iii) train the ensemble using randomized initialization of model parameters to increase variety in the ensemble~\cite{Lakshminarayanan:2017}. Many common loss functions, such as cross-entropy loss, are strictly proper scoring rules and can therefore be used in the deep ensemble framework. In practice, adversarial training is often omitted if improved robustness is not strictly necessary. \textcite{Lakshminarayanan:2017} also demonstrate that the model has the attractive property of decreasing its certainty of prediction in out-of-distribution examples, which was demonstrated using an ensemble trained on the MNIST dataset on examples from the NotMNIST dataset which contains letters instead of digits. It has later been verified that DEs are the SotA for UQ on OOD data \cite{Ovadia:2019, Ashukha:2020}. For these reasons, and that training time is not critically limited in the same way as computational capacity upon test-time inference, we chose DEs as the framework for generating our ensemble. It should be noted that we are not the first to attempt leveraging ensembles while keeping the test-time inference time down. Wortsman \etal~\cite{wortsman2022model} propose to average the weights of ensemble members into a single model. Havasi \etal~\cite{havasi2020training} propose to divide a given neural network into sub-networks, which essentially constitute an ensemble.

\parsection{Monte Carlo dropout}
Dropout was first introduced by \textcite{Srivastava:2014} as a regularization measure during training to limit overfitting and increase the generalizability of the learned representation. With dropout, each neuron is turned off at random during training according to a pre-specified probability, or dropout rate, $p$. This helps the network not to overfit, and therefore generalize better, as it has to create a more robust representation when any neuron can be dropped at any time. Recognizing that using an ensemble of a set of models is usually beneficial for model performance, \textcite{Srivastava:2014} show that using dropout during inference is equivalent to sampling from an exponential set of possible smaller models, which yields higher overall performance. \textcite{Gal:2016} later showed that performing a number of forward passes through a model with dropout enabled and averaging the results can be seen as a Bayesian approximation. They chose to call this MC-dropout and claimed that it enables superior uncertainty estimation performance in both regression and classification tasks compared to vanilla models. Of note is that since the introduction of MC-dropout, \textcite{Lakshminarayanan:2017, Ovadia:2019, Ashukha:2020}, have all claimed that deep ensembles are superior in uncertainty quantification. However, MC-dropout remains widely used due to its simple implementation and general improvement of performance compared to vanilla single models. This makes it a relevant mode of comparison for our proposed model.

\parsection{Traffic sign recognition}
We evaluate the proposed method on Traffic Sign Recognition (TSR), a field with a decades-long history of development \cite{Stallkamp:2012}. Lately, TSR has become an important part of AD systems, and high and reliable performance is important for safety. This makes it an interesting application for DESOT. 

There are two main subproblems of TSR, traffic sign detection and traffic sign classification \cite{Mathias:2013}. This research concerns itself with the latter and assumes that regions of interest in the image have already been identified by another system (in this specific case, human annotators) earlier in the ML pipeline. The domain is characterized by a large number of classes with an imbalanced class distribution. Additionally, variations in illumination, perspective, and occlusions are common \cite{Stallkamp:2012}, making the problem distinctly long-tailed. Furthermore, many of the classes are very similar in shape and color, but with important differences in meaning, such as speed limit signs. Deep learning has recently started revolutionizing this domain, with many models achieving accuracies of over 95\% in research settings \cite{Magnussen:2020}. This means that any differences in predictive performance between the models tested are likely to be small in absolute terms, and performance benefits might instead lie in the performance of the models on difficult examples such as short sequences or obscured scenes.

\begin{figure*}[t]
    \centering
    \begin{subfigure}{0.7\linewidth}
      \centering
      \includegraphics[width=1\hsize]{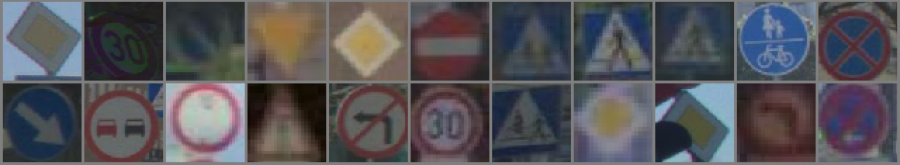}
      \caption{Single frames.}
      \label{fig:training_examples}
    \end{subfigure}
    \\
    \begin{subfigure}{0.7\linewidth}
      \centering
      \includegraphics[width=1\hsize]{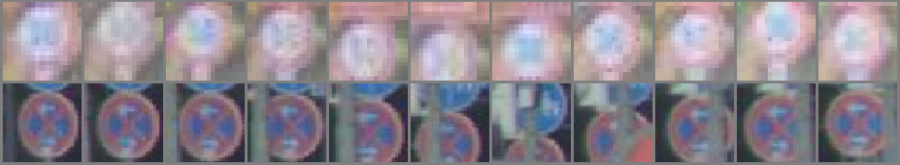}
      \caption{Sequences dataset.}
      \label{fig:test_examples}
    \end{subfigure}
    \caption{Examples of training and validation data. Training data is cropped from ZOD, and sequences are created by additional tracking across time using adjacent frames.}
    \label{fig:data_examples}
\end{figure*}

\section{Deep Ensembles Spread over Time}

In this section, we introduce Deep Ensembles Spread Over Time (DESOT), a method to obtain the benefits of a Deep Ensemble (DE) without additional computational cost. In many real-world scenarios, the input data is a sequence of frames. In the literature, it is then common to apply a single model and average the predictions~\cite{Brissman:2023}. This strategy is illustrated in Fig.~\ref{fig:traditional_singlemodel}. It is well-known that replacing the single model with a DE improves both predictive- and uncertainty estimation performance (Fig.~\ref{fig:traditional_ensemble}). Unfortunately, this multiplies the computational cost with a factor $M$, where $M$ is the number of ensemble members. DESOT (Fig.~\ref{fig:DESOT_visualization}) instead divides the ensemble over time, running one ensemble member per frame, and thus has the same computational cost as a single model. We hypothesize that DESOT still provides, despite the cheap computational cost, the benefits of DEs.

\subsection{Image Classification on Sequences}
We will now introduce the application of a single model or DE to \emph{image classification on sequences}~\footnote{Note that image classification on sequences is different from video classification, as the former does not need to model the dynamics of the scene.}. First, both the ensemble fusion block and the time fusion block are simple averaging operations. This is how \textcite{Lakshminarayanan:2017} implement ensemble fusion for deep ensembles in the classification setting. Now, define a sequence $\mathbf{x} \in \mathbb{R}^{T \times H \times W \times 3}$ as $T$ distinct still images, each with a height of $H$ pixels, a width of $W$ pixels, and three separate color channels. We will consider classification problems where a model should produce a categorical output distribution across $C$ classes for such a sequence $\mathbf{x}$. Assume there is a set of $M$ different neural networks, each of which can conduct this classification. A single model $m \in \{1, ..., M\}$ produces a categorical output distribution $\mathbf{p}_m(x^t)$ for each single image $x^t$ at time step $t \in \{1, ..., T\}$ in the sequence $\mathbf{x}$. Then, the final output distribution for model $m$ is defined as

\begin{equation}\label{eq:SM_sequence_prediction}
    \mathbf{p}_m(\mathbf{x}) = \frac{1}{T}\sum^T_{t=1} \mathbf{p}_m(x^t)\enspace,
\end{equation} which is the element-wise (class-wise) average across the output distributions for each image in the sequence at different time steps. This setup is what we will refer to as a single model (SM).

Now imagine that all $M$ models are used for classifying each image in the sequence, such that the final output distribution for the sequence is 

\begin{equation}\label{eq:DE_sequence_prediction}
    \mathbf{p}_{\text{DE}}(\mathbf{x}) = \frac{1}{M} \sum^M_{m=1} \mathbf{p}_m(\mathbf{x}) = \frac{1}{MT} \sum^M_{m=1} \sum^T_{t=1} \mathbf{p}_m(x^t)\enspace,
\end{equation} which is how we choose to apply deep ensembles (DEs) \cite{Lakshminarayanan:2017} to sequences -- averaged across the images of the sequence $\mathbf{x}$. DE$_M$ will be used to denote an $M$-member deep ensemble. 

\subsection{DESOT}\label{subsec:DESOT}
Our proposed method, which we call DESOT, instead uses a single model $m \in \{1, ..., M\}$ for each image $x^t$, $t \in \{1, ..., T\}$, to produce a categorical output distribution, but the models are alternated such that any given model $m$ is used on average $T/M$ times for a certain sequence of $T$ images. Analogously to the notation used for DEs, DESOT$_M$ will denote an $M$-member deep ensemble spread over time.
We let $\sigma(t)$ denote the (repeated) mapping between each time step $t$ and one of the ensemble members $m \in \{1, ..., M\}$, and write the final output distribution produced by DESOT across time steps, ensemble members, and classes as
\begin{equation}\label{eq:DESOT_sequence_prediction}
    \mathbf{p}_{\text{DESOT}}(\mathbf{x}) = 
\frac{1}{T} \sum^{T}_{t=1} \mathbf{p}_{\sigma(t)}(x^t)\enspace,
\end{equation} 
As is clear from this definition, let us again stress that the DESOT method can only be applied to sequences since it fundamentally relies on alternating the ensemble member in use between neighboring frames. If DESOT was to be used on a sequence length of one, the model would be equivalent to a standard single model. 

In the real world, one would have a continuous stream of frames from the cameras of the car. In such a case, a set window size might be used such that equations \ref{eq:SM_sequence_prediction}-\ref{eq:DESOT_sequence_prediction} each constitute a moving average across some previous images. Due to the shortness of the sequences we use in this paper, which are only 11 frames, we have chosen to use the average across the entire sequence for a model. 

\section{DESOT for Traffic Sign Recognition}
We apply DESOT to traffic sign classification, a challenging image classification problem that is a cornerstone to autonomous driving. While traffic sign classification is oftentimes studied as a single-frame problem, autonomous vehicles will in practice obtain a sequence of frames.

\subsection{Data}
We use the Zenseact Open Dataset (ZOD)~\cite{ZOD:2023}, a multi-modal dataset collected by the autonomous driving software company Zenseact. The data has been collected across a number of countries in Europe. For this work, we use the single-frame data in ZOD, which has high-quality annotations for 446k unique signs. Using the annotations, the traffic signs are cropped and saved separately. 
The majority of these signs are withheld for single-frame training and validation. Due to the lack of temporal sign annotations, 30k randomly selected signs are extended into sequences of crops using an off-the-shelf tracker on the 10 preceding frames, sampled at 15 Hz. These sequences are only used for testing. See \autoref{fig:data_examples} for samples from the datasets. 

\subsection{Architecture and training}
We use Resnet18 as the base for all our models tested. Note however that, just like DEs, our approach is architecture agnostic. In addition to the single model and traditional DE, we compare DESOT to an MC-dropout implementation. We use a dropout layer after each non-linearity with a dropout rate of 0.2. The dropout-version is run once on each frame in a sequence and then aggregated by averaging in the same way as for the other models. All models are trained for 30 epochs using the AdamW optimizer \cite{Loshchilov:2017} with a learning rate of 0.0005. Cosine annealing is used for more stable convergence. A batch size of 256 is used. Classes with fewer than 10 occurrences in the training- and validation datasets are omitted from training and evaluation, as well as crops smaller than 16 pixels along any dimension. In line with the definition of deep ensembles by \textcite{Lakshminarayanan:2017}, all models are trained independently using a proper scoring rule. We use cross-entropy loss. 

For comparisons on uncertainty quantification (UQ), all models are temperature scaled. We use the procedure first introduced by \textcite{Guo:2017} where a separate validation set is used to find the temperature that optimizes negative log-likelihood. The validation set in question is a separate subset of single frames from ZOD. For the deep ensemble, a joint temperature scaling scheme is employed where the ensemble is treated as a single model with one temperature parameter. This was found to be superior to creating an ensemble of single models of perfect temperature. This finding agrees with research by \textcite{Rahaman:2021}. We adopt the same temperature scaling for DESOT.

\section{Results}
Our study comprises an analysis of predictive performance (Sec.~\ref{sec:result-pred}) and out-of-distribution uncertainty quantification (Sec.~\ref{sec:result-ood}). The latter is divided into two parts, focusing firstly on real out-of-distribution examples and secondly on systematically increasing the severity of augmentations for in-distribution examples.

\begin{table*}[t]
    \centering
    \caption{Performance of each strategy on the ZOD sequence dataset. The results include $\pm$ one standard deviation of performance over 5 runs. All models are temperature scaled. Deep Ensemble (DE) improves over the single model (SM) in terms of predictive- and uncertainty estimation performance, but by multiplying the computational cost with the number of ensemble members. Our DESOT, in contrast, obtains the DE benefits while avoiding the added computational cost.}
    \resizebox{0.7\textwidth}{!}{\begin{tabular}{c c c c c}
        \hline
        \textbf{Model} & \textbf{Accuracy} $\uparrow$ & \textbf{F1-score} $\uparrow$ & \textbf{Brier reliability} $\downarrow$ & \textbf{ECE} $\downarrow$\\ \hline
        SM (single frame) & 94.85 $\pm$ 0.06 & 70.42 $\pm$ 1.27 & \textbf{0.0069 $\pm$ 0.0007} & \textbf{0.21 $\pm$ 0.06} \\
        SM      & 97.34 $\pm$ 0.06 & 81.12 $\pm$ 1.75 & 0.0124 $\pm$ 0.0003 & 2.88 $\pm$ 0.06 \\
        DESOT$_5$      & 97.60 $\pm$ 0.03 & \textbf{83.26 $\pm$ 0.93} & 0.0108 $\pm$ 0.0002 & 2.25 $\pm$ 0.05 \\
        DE$_5$         & \textbf{97.64 $\pm$ 0.01} & 82.73 $\pm$ 1.01 & 0.0108 $\pm$ 0.0002 & 2.25 $\pm$ 0.05 \\
        MC-dropout     & 97.10 $\pm$ 0.09 & 76.79 $\pm$ 2.71 & 0.0165 $\pm$ 0.0004 & 4.03 $\pm$ 0.07\\
        \hline
    \end{tabular}}
    \label{tab:predictive_performance}
    \vspace{-4mm}
\end{table*}

\subsection{Predictive performance}\label{sec:result-pred}
We assessed the predictive performance of various models on the sequence dataset. Throughout the later epochs, all models achieved high accuracy, yet DESOT$_5$ and DE$_5$ particularly outperformed their counterparts in early epochs. Importantly, DESOT matched the performance of traditional DEs, with both models slightly surpassing a single model with temporal fusion (SM). Comprehensive performance details are documented in \autoref{tab:predictive_performance}. When assessed via F1-score, DESOT maintained parity with DEs, creating a wider gap to SM looking at averages across runs. However, do note that variance also increases compared to accuracy. The inclusion of MC-dropout in the single model significantly diminished its performance. Notably, all models, including SM, significantly outperformed our baseline -- a single-frame model operating on a randomly selected sequence frame without temporal fusion.

We also evaluated the calibration of each model using Brier reliability and Expected Calibration Error (ECE). Notably, a considerable gap emerged between the single model operating on a single frame and the one fused over the entire sequence. This disparity could be linked to the single-frame temperature scaling procedure used, and the tendency of averaging predictions over multiple images to reduce overall confidence, thereby negatively impacting calibration. However, both DESOT and DE substantially improved sequence calibration, even though they still fell significantly short of the single-frame calibration. These findings suggest the potential value of a sequence-aware fine-tuning step to create better-calibrated models for sequences.

An essential aspect of traffic sign recognition, and by extension, classification, is the long-tailed distribution of classes. The fact that high performance on all classes is important accentuates the significance of performance on rare classes within the aggregated performance metrics. To evaluate the models' performance on these rare classes, we removed classes with more than 500 samples in the training dataset, leaving us with 625 sequences to evaluate performance on. The predictive performance results for this filtered dataset can be found in \autoref{tab:predictive_performance_small}. Due to the smaller dataset (both in training and validation) the results are significantly more noisy. Nevertheless, the results indicate that both ensembling approaches improve performance for these difficult and rare cases.

Overall, our method notably outperforms a single model with a similar computational footprint, while matching the performance of DE$_5$, which has a five-fold larger computational footprint. This superior performance could be attributed to the ensemble members collectively offering a more expressive representation of potential traffic signs than a single model. The distinct advantage of DESOTs lies in their capacity to leverage this richer representation while minimizing computational demands.

\begin{table}[t]
    \centering
    \caption{Predictive performance on a minority class version of the ZOD sequence dataset. Without extra computational cost beyond the single-model baseline, DESOT substantially outperform the single model. In contrast to the full dataset, the single-frame-single-model strategy yields competitive performance.}
    \small
    \resizebox{0.9\linewidth}{!}{\begin{tabular}{c c c}
        \hline
        \textbf{Model} & \textbf{Accuracy} $\uparrow$ & \textbf{F1-score} $\uparrow$ \\ \hline
        SM (single frame) & 87.42 $\pm$ 0.97          & 46.49 $\pm$ 3.03 \\
        SM                & 87.70 $\pm$ 0.58          & 43.65 $\pm$ 2.84 \\
        DESOT$_5$         & \textbf{89.53 $\pm$ 0.71} & \textbf{47.53 $\pm$ 1.58} \\
        DE$_5$            & 89.41 $\pm$ 0.38          & 46.57 $\pm$ 2.19 \\
        MC-dropout        & 85.52 $\pm$ 0.87          & 36.50 $\pm$ 0.71 \\
        \hline
    \end{tabular}}
    \label{tab:predictive_performance_small}
    \vspace{-5mm}
\end{table}

\subsection{Out-of-distribution uncertainty quantification performance}\label{sec:result-ood}
In the real world, ML systems often encounter data absent from the training set, known as Out-of-Distribution (OOD) data. Thus, it's critical to evaluate how a system performs under such circumstances. Particularly, since machine learning systems can fail silently, confidently misclassifying OOD examples \cite{Amodei:2016}, we aim to characterize the behavior of various systems when confronted with OOD data. This is achieved by testing gradual OODness via augmentations, akin to \textcite{Ovadia:2019}, and by assessing performance on completely unseen data, as per \textcite{Lakshminarayanan:2017}. This enables us to simulate common scenarios (e.g., rotated signs or changing lighting) and identify changes in output distribution for OOD detection.

\subsubsection{Complete out-of-distribution data}
The Zenseact Open Dataset includes a class named \texttt{NotListed}, representing traffic signs not included in any other class, such as destination signs and rare special signs. This class serves as an unseen set of mixed traffic signs for OOD data testing. Since the models were not trained on this OOD data, previously used metrics like accuracy, F1-score, Brier score, and ECE aren't applicable. Instead, we use entropy as a metric to quantify if the models 'know what they don't know'. Higher entropy of the output distribution suggests higher model uncertainty.

\begin{figure*}[t]
    \centering
      \includegraphics[width=.9\hsize]{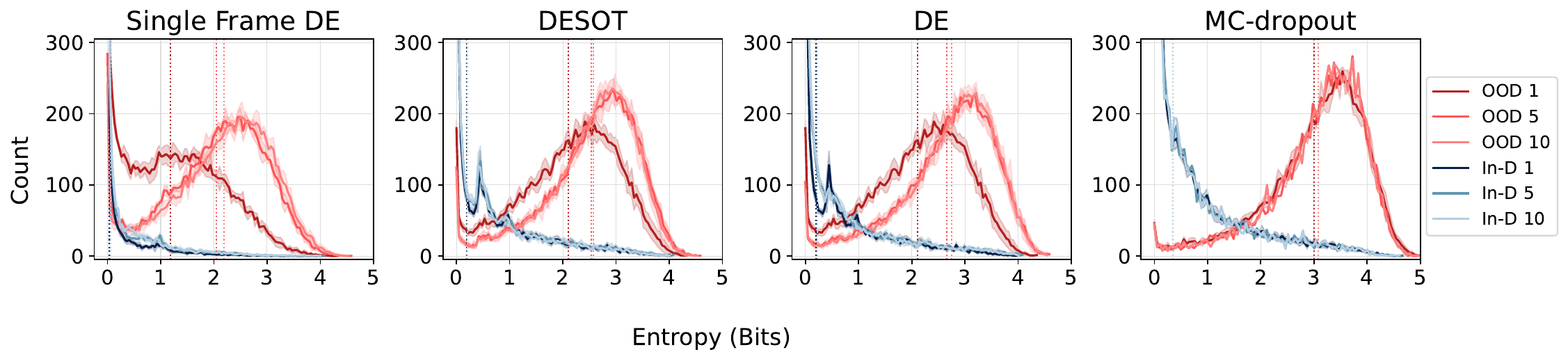}
    \vspace{-3mm}
    \caption{Histogram of the entropy for OOD data (red) and in-distribution data (blue) for the ZOD sequence dataset. We run various ensemble sizes $M\in \{1,5,10\}$, which are differentiated by color shade. Again, note that DE$_1$ and DESOT$_1$ are special cases that are equivalent to a single model. The vertical dashed lines are the mean entropy for the model of the same color. Like standard deep ensembles, DESOT increases its entropy on OOD data as additional ensemble members are added.}
    \label{fig:entropy_ood}
    \vspace{-3mm}
\end{figure*}

\begin{figure*}[t]
    \centering
    \includegraphics[width=0.7\textwidth]{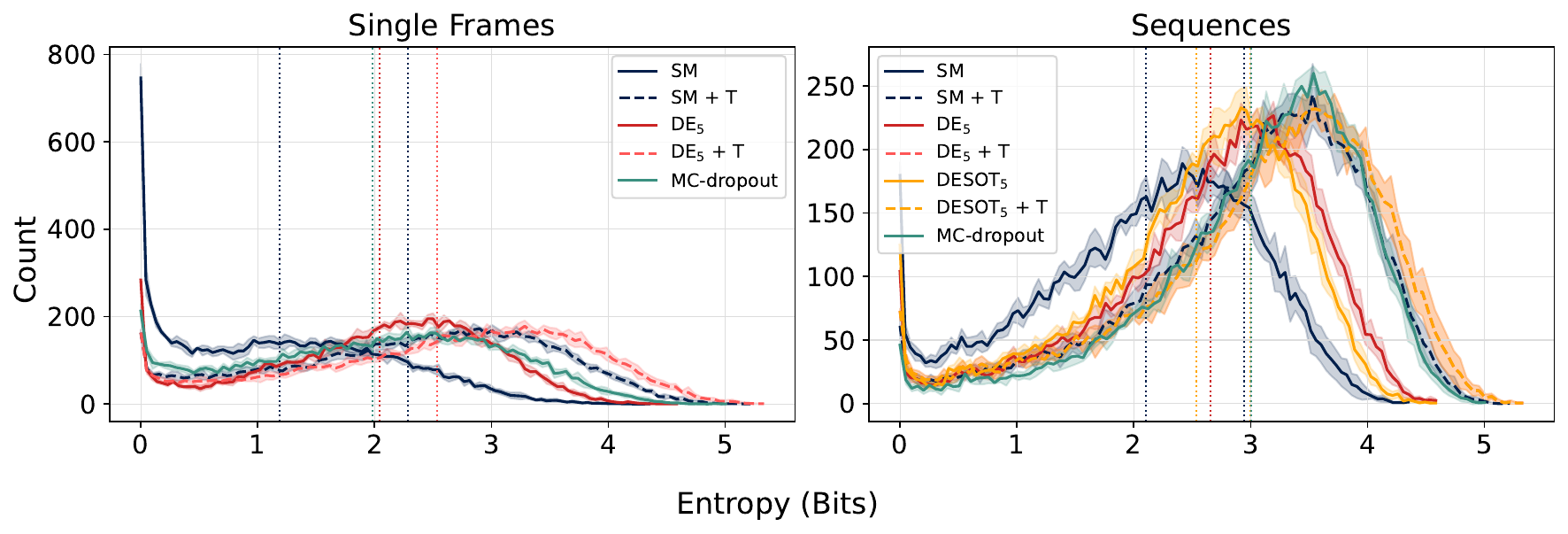}
    \vspace{-2mm}
    \caption{Histogram of the entropy for OOD data for the single-frame test dataset (left) and the sequences dataset (right). The $+T$ suffix denotes temperature scaling. Similar to deep ensembles, DESOT increases the entropy for OOD data, facilitating OOD detection.}
    \label{fig:entropy_ood_ensembles}
    \vspace{-3mm}
\end{figure*}

The results are illustrated via graphs showcasing the entropy scores of different models. The effect of varying ensemble sizes can be seen in \autoref{fig:entropy_ood}, whereas different approaches are more easily compared in \autoref{fig:entropy_ood_ensembles}. As expected, all models exhibit a drastic increase in entropy for OOD data compared to in-distribution data. The single-frame model shows limited adjustment to its uncertainty on OOD data. On the contrary, single-frame DE$_5$ and DE$_{10}$ demonstrate a substantial entropy increase on OOD data, which is consistent with previous research \cite{Lakshminarayanan:2017, Ashukha:2020}. Interestingly, the increase from 5 to 10 ensemble members is relatively minor, suggesting a saturation effect.

When evaluated over a sequence, the entropies of DEs and DESOTs exhibit remarkable similarity, with a slight advantage towards DEs. Surprisingly, the entropy distribution of single models, when aggregated over a sequence, nearly mirrors that of single-frame ensembles. Further, MC-dropout also escalates its entropy in a manner consistent with DEs and DESOTs, contradicting the findings of \textcite{Lakshminarayanan:2017}. However, we must emphasize that the adoption of MC-dropout results in degraded predictive performance, as highlighted in our prior results.

Next, we used a simple thresholding strategy on the output entropy to test each model's OOD detection potential, rewarding methods that clearly separate in- and out-of-distribution entropy distributions. We fitted a threshold to each model using half of the available sequences to detect in- or out-of-distribution samples. The other half of the sequences dataset was then used to evaluate OOD detection performance for each model, with results presented in \autoref{tab:ood_threshold}, including the entropy threshold value and metrics such as accuracy, precision, recall, and F1-score.

While the optimal threshold value varied widely between models, the OOD detection performance of all models was relatively high. In a standard setting, DESOT and DE perform closely and clearly outperform the single model across all metrics. The introduction of temperature scaling, typically used to enhance the calibration of classification models, alters this picture. It marginally impacts the ensembles but dramatically improves the single model's performance, almost entirely closing the gap with the ensembles. Surprisingly ensembles usually outdo temperature scaling in OOD detection. As our focus is the comparison between DESOTs and DEs, we do not further explore temperature scaling. 

We also acknowledge insights from other research \cite{Kirsch:2021}, highlighting potential pitfalls of this simple experiment, such as the risk of misidentifying ambiguous in-distribution samples as OOD -- a problem we also encountered. Despite this, these tests offer a fundamental benchmark for more sophisticated OOD detection strategies, thereby illustrating the effectiveness of even such a straightforward method.

\begin{table}[t]
    \centering
    \caption{Results from applying an entropy threshold for OOD detection on the sequences dataset. Note that DEs and DESOTs perform the best out of all models. When subject to temperature scaling (rows with $+T$ suffix), the difference in performance decreases.}
    \vspace{-2mm}
    \label{tab:ood_threshold}
    \resizebox{\linewidth}{!}{\begin{tabular}{l c c c c c}
    \hline
    & Threshold & Accuracy & Precision & Recall & F1 \\ \hline
    SM & 0.709 & 0.895 & 0.811 & 0.905 & 0.855 \\
    DESOT$_5$ & 1.048 & \textbf{0.919} & 0.852 & \textbf{0.922} & \textbf{0.886} \\
    DE$_5$ & 1.071 & \textbf{0.919} & \textbf{0.856} & 0.919 & \textbf{0.886} \\
    MC-dropout                & 1.428 & 0.910 & 0.845 & 0.900 & 0.872 \\
    \hline
    SM + T & 1.463 & 0.916 & \textbf{0.856} & 0.906 & 0.880 \\
    DESOT$_5$ + T & 1.143 & 0.919 & 0.850 & \textbf{0.928} & 0.887 \\
    DE$_5$ + T & 1.178 & \textbf{0.920} & 0.853 & 0.925 & \textbf{0.888} \\
    MC-dropout + T & 1.610 & 0.914 & 0.842 & 0.921 & 0.880 \\
    \hline
    \end{tabular}}
    \vspace{-4mm}
\end{table}

\subsubsection{Shifted out-of-distribution data}
We generate increasingly out-of-distribution data by escalating the severity of augmentations, inspired by the works of \textcite{Hendrycks:2019} and \textcite{Ovadia:2019}. This experiment involved six distinct augmentations, each with a specific intensity range. As we increase the augmentation's intensity, we anticipate a decrease in model accuracy. Ideally, the models should adjust their certainty to correspond with this accuracy drop, thereby maintaining calibration. Reliable model uncertainty estimates offer actionable insights to downstream users.

However, when augmentation intensities are high, calibration may not provide a meaningful measure of uncertainty quantification. Since accurate sequence classification often becomes challenging under intense augmentations, it's more crucial to identify such samples as out-of-distribution rather than obtaining a well-calibrated class distribution, as the ground truth class loses its relevance. Hence, we have also recorded the mean entropy across the dataset for each model and augmentation severity.

\begin{figure*}[t]
    \centering
    \includegraphics[width=1.0\hsize]{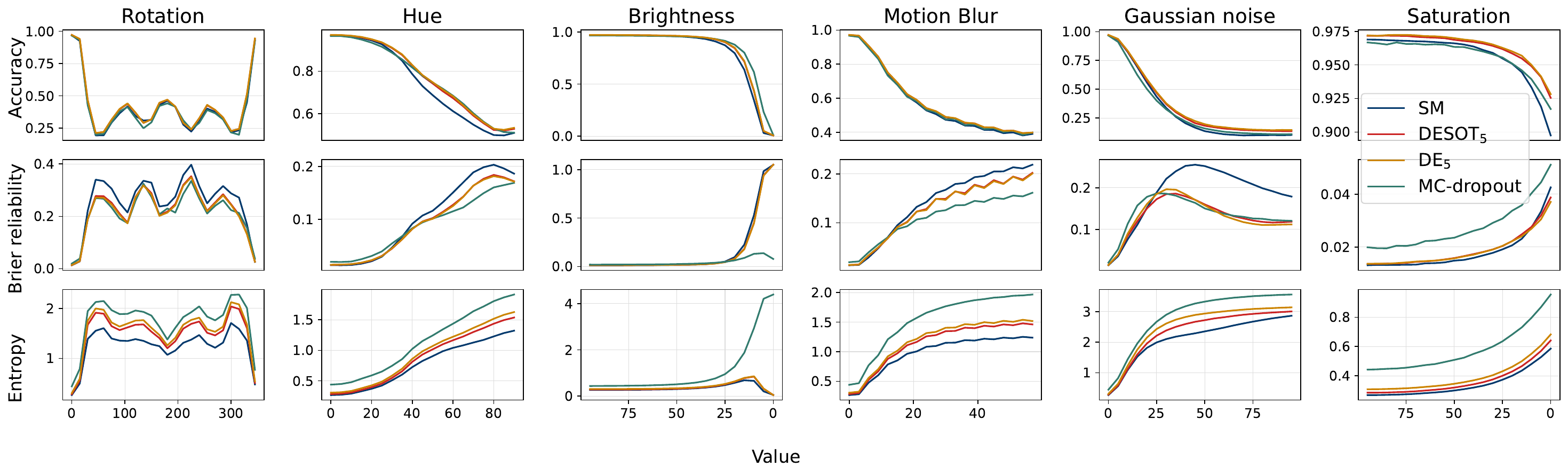}
    \vspace{-4mm}
    \caption{Uncertainty quantification performance for each model on augmented data of increasing intensity. The performance is measured in accuracy, Brier reliability, and mean entropy. Tested on the sequences dataset. All approaches tend to increase their entropy as augmentations become stronger. For some augmentations, the single model without temperature scaling and the MC-dropout increase entropy much more or much less than the other approaches.}
    \label{fig:augmentations}
    \vspace{-4mm}
\end{figure*}

\autoref{fig:augmentations} illustrates how various augmentations affect accuracy, Brier reliability~\cite{DeGroot:1983,Siegert:2017}, and mean entropy of models in sequence settings at different intensity levels. Lower Brier reliability indicates better model calibration, while higher mean entropy suggests increased model uncertainty. For brevity, we omit results for the single-frame setting.

Observing the graphs, we notice a trend where a decrease in model accuracy corresponds with an increase in the Brier reliability score. This suggests that models become less calibrated as the data turns more out-of-distribution. Furthermore, entropy appears to rise with augmentation intensity, inversely correlating with accuracy. The strange trends in the rotation augmentation graphs are attributed to many signs having less than $360\degree$ rotational symmetry.

Examining the results from augmented OOD data, it appears that all models show comparable robustness to the augmentations in terms of predictive performance, deteriorating at similar rates for increased intensities. However, there's a greater variation in model calibration upon augmentation. Single models struggle to maintain calibration at higher augmentation intensities, particularly for rotations, hue, motion blur, and Gaussian noise, a phenomenon also observed by \textcite{Ovadia:2019}. As in the earlier OOD study, MC-dropout delivers convincing robustness, albeit at the expense of predictive performance. It is important to note that our study and that of \textcite{Ovadia:2019} are performed on distinct datasets -- our more complex sequence data versus their simpler MNIST dataset.

In general, DEs and DESOTs seem to offer similar calibration performances, measured in Brier reliability, for a given augmentation intensity, outperforming single models. Post temperature scaling, the disparity between the ensembling methods and single models is not as pronounced. An exception is Gaussian noise corruption, where single models distinctly fail to maintain calibration compared to others.

\section{Conclusions}

We have introduced Deep Ensembles Spread Over Time (DESOT), a novel approach that distributes ensemble members across a sequence, achieving the predictive power and out-of-distribution robustness of running a full deep ensemble at each time step, with the computational cost of a single model. We extensively demonstrate the viability of this approach on the task of traffic sign classification, a highly relevant task where misinterpretations can lead to catastrophic outcomes and out-of-distribution signs are prevalent.

Looking ahead, we see exciting potential for DESOTs in more complex scenarios, such as 3D object detection where one could use a Kalman Filter for temporal fusion. We believe that our work opens up an avenue for new research on high-performing resource-efficient models, by demonstrating that ensembles can indeed be spread over time.

\parsection{Limitations} The proposed strategy, DESOT, is applicable only for sequence processing problems. Although, in theory, the computational resources should match those required for a single model, it's practical to anticipate that at least two members of the ensemble might need concurrent loading into memory. This scenario could potentially double the memory requirements. Moreover, while the improvements of DESOT (and deep ensembles) over SM are significant, we were surprised to see the large benefits of SM over the single-frame-single-model. A similar observation can be made for the out-of-distribution detection, where the effect of temperature scaling almost equals that of DESOT or deep ensembles. Though, combining DESOT or DEs with temperature scaling provides the best performance. We also note that the model calibration is actually better on a single frame than it is for multiple frames. One potential future research direction is thus to investigate how temperature scaling is best applied to the image classification on sequences problem.

\parsection{Acknowledgements} This work was partially supported by the Wallenberg AI, Autonomous Systems, and Software Programme (WASP).
{\small
\printbibliography
}

\end{document}